\documentclass{svproc}
\usepackage{microtype}
\usepackage{url}
\usepackage{lmodern}
\usepackage[T1]{fontenc}
\usepackage[utf8]{inputenc}
\usepackage{graphicx}
\usepackage{amsmath}
\usepackage{xcolor}
\usepackage{colortbl}
\usepackage{tikz}
\usepackage{booktabs}
\usepackage{multirow}
\usepackage{authblk}
\usepackage{hyperref}
\usepackage{float}
\usetikzlibrary{arrows, positioning}

\pdfoutput=1

\begin{document}

\mainmatter
\title{Optimized Quran Passage Retrieval Using an Expanded QA Dataset and Fine-Tuned Language Models}
\titlerunning{Quran QA}

\author{
    Mohamed Basem \and 
    Islam Oshallah \and 
    Baraa Hikal \and
    Ali Hamdi \and 
    Ammar Mohamed
}
\institute{
Faculty of Computer Science, MSA University, Egypt \\
\email{\{mohamed.basem1, islam.abdulhakeem, baraa.moaweya, ahamdi, ammohammed\}@msa.edu.eg}
}

\maketitle
\begin{sloppypar}

\begin{abstract}
Understanding the deep meanings of the Qur’an and the bridge the language gap between modern standard Arabic and classical Arabic is essential to improve the question-and-answer system for the Holy Qur’an. The Qur’an QA 2023 shared task dataset had limited number of questions with weak model retrieval. To address this challenge, this work was done to update the original dataset and improve the model accuracy. The original dataset which contains 251 questions was reviewed and expanded to 629 questions with questions diversification and reformulation, leading to a comprehensive set of 1895 categorized into single-answer, multi-answer, and zero-answer types. Extensive experiments fine-tuned transformer models, including AraBERT, RoBERTa, CAMeLBERT, AraELECTRA, and BERT. The paper best model, \textbf{AraBERT-base}, achieved a \textbf{MAP@10 of 0.36} and \textbf{MRR of 0.59,} representing improvements of 63\% and 59\%, respectively, compared to the baseline scores (MAP@10: 0.22, MRR: 0.37). Additionally, the dataset expansion led to improvements in handling” no answer” cases, with the proposed approach achieving a \textbf{75\% success rate} for such instances, compared to the baseline’s 25\%. These results demonstrate the effect of dataset improvement and model architecture optimization in increasing the performance of QA systems for Holy Qur’an, with higher accuracy, recall, and precision.

\keywords{Quran Question Answering,
Passage Retrieval,
Modern Standard Arabic}
\end{abstract} 
\section{Introduction}
In context of expanding the number of Muslim populations worldwide (2.04 billion in 2024), there Is      a growing demand for understanding the holy Quran through a reliable question-answering (QA) system capable of providing exact explanations and answers from the Qur’an \cite{elkomy2023tce}. In recent years, the Qur’an QA 2023 shared task dataset has highlighted the complexity of this task, as previous methods have shown limited accuracy in retrieving relevant Qur’anic verses \cite{mahmoudi2023multi}. Many traditional QA models face difficulties with the linguistic nuances of
Classical Arabic and the specificity required in Holy Qur’an \cite{essam2024survey}. the study of this work builds upon this challenge by expanding the existing dataset and employing advanced language models to improve retrieval accuracy \cite{elkomy2023tce}.
The contributions in this paper are as follows:
\begin{itemize}
    \item \textbf{Dataset Manipulation and Expansion}: We expanded the original QA dataset by generating new questions through rephrasing and categorization, resulting in a significantly larger and more diverse set of 1895 questions.
    \item \textbf{Accurate Language Model Fine-Tuning}: We fine-tuned multiple transformer models on the expanded dataset, achieving notable improvements in passage retrieval accuracy, particularly with the AraBERT-large model.
\end{itemize}
The layout of this article is outlined in the following manner: Section 1 explore into the reasons and importance of improving Qur’anic question-answering systems. Section 2 gives a summary of previous research, outlining major progress and obstacles in the area. In Section 3, The research methodology is described, including data collection, expanding the dataset, cleaning the data, and finetuning the language model. The setup utilized to assess model performance is explained in Section 4’s experimental design. Section 5 highlights and discusses the outcomes of experiments are presented, showcasing the efficiency of the method. In conclusion, Section 6 of the paper presents suggestions for upcoming studies.

\section{Related Work}
The field of question answering (QA) for Holy Qur’an, specifically Qur’anic passage retrieval, has attracted a lot of attention from researchers because of   the distinct linguistic and contextual difficulties presented by the Qur’an \cite{malhas2023phd}. The task requires accurately retrieving relevant verses to answer both factoid and non-factoid questions, often requiring systems to bridge the linguistic gap be- tween Modern Standard Arabic (MSA) and Classical Arabic \cite{malhas2023overview}. 
Additionally, systems must be able of recognizing questions that have no answers within the Qur’anic text, thereby requiring strong mechanisms for zero-answer scenarios \cite{sardar2017reading}. The development of effective QA systems for Holy Qur’an remains a challenging endeavor, as the richness of the Arabic language and the need for contextual understanding demand advanced modeling techniques \cite{hillman2019qa}.

Transformer-based language models, such as AraBERT, CAMeLBERT, and AraELECTRA, have shown promise in handling Arabic language tasks \cite{aljamel2024fine}. However, existing models often struggle with limitations stemming from insufficient and imbalanced training datasets, which impact the models’ ability to generalize effectively to unseen queries \cite{liu2023imbalanced,hamdi2016review,hamdi2018clasenti}. The Qur’an QA 2023 shared Task, for example, highlighted the necessity of utilizing external resources and data augmentation strategies to improve performance. Sarhan and Elkomy \cite{elkomy2023tce} addressed these challenges by leveraging ensemble learning techniques, combining dual-encoder and cross-encoder architectures, and employing transfer learning on external datasets such as TyDI-QA and tafseer. Their ensemble strategy improved prediction stability and effectiveness, achieving a MAP score of 25.05\% for passage retrieval. Despite these advancements, their study underscored the persistent need for dataset expansion and more robust fine-tuning approaches to tackle the linguistic complexities of Qur’anic QA tasks \cite{elkomy2023tce}.

The work of Alawwad et al \cite{alawwad2023ahjl} focused on enhancing Qur’anic passage retrieval through the use of pre-trained models fine-tuned on specialized datasets like Tafseer and TyDI-QA \cite{clark2020tydi}. They merged thresholding mechanisms to manage unanswerable questions and demonstrated the benefits of ensemble learning for improving performance in low-resource settings. Although their approach showed considerable success, achieving improved performance metrics, it also pointed to the critical role of dataset quality and quantity in achieving consistent model accuracy. The limitations of existing datasets and the models’ dependency on external resources highlight the need for further dataset expansion and optimization strategies.
Mahmoudi et al \cite{mahmoudi2023multi} proposed a multi-task transfer learning approach that employs models like AraElectra and AraBERT, utilizing both unsupervised and supervised fine-tuning to adapt to the Qur’anic context. They implemented tech-niques like TSDAE and SimCSE to produce top-notch sentence embeddings, greatly improving the models’ ability in reading comprehension and passage retrieval. Their study showed that custom sentence embeddings and thorough model fine-tuning can result in actual enhancements. Nonetheless, the study also emphasized that further advancements require larger and more diverse datasets to fully Understand the complexities of the language and context of the Quran. This work builds on these prior efforts by addressing the pressing need for dataset expansion and more effective model fine-tuning \cite{zheng2023learn}. A significant increase was achieved the size of the Qur’an QA dataset from 251 to 1895 questions, employing strategic question rephrasing to improve data diversity and model robustness \cite{sun2020mixup}. The proposed approach involves extensive experimentation with multiple transformer- based models, including AraBERT, CAMeLBERT, and AraElectra, fine-tuning them on this enriched dataset. By doing so, the aim of this work is to enhance the models’ ability to handle complex queries, manage zero-answer cases, and improve overall passage retrieval accuracy, contributing to the advancement of QA systems for Holy Qur’an \cite{alawwad2023ahjl}.
 
\section{Research Methodology}
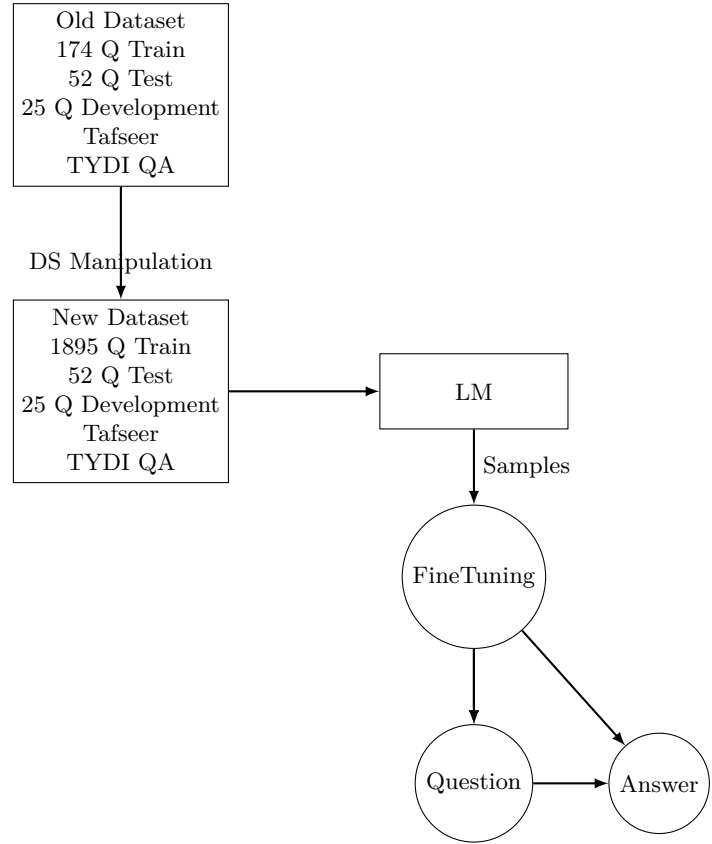
\begin{figure}[H]
    \centering
    \begin{tikzpicture}[
        node distance=2cm,
        box/.style={draw, rectangle, minimum width=2.5cm, minimum height=1cm, align=center},
        circlebox/.style={draw, circle, minimum size=1cm, align=center},
        arrow/.style={-latex, thick}
    ]

    \node[box] (QA) {Old Dataset\\ 174 Q Train\\ 52 Q Test\\ 25 Q Development \\ Tafseer\\ TYDI QA};
    \node[box, below=1.5cm of QA] (NewDS) {New Dataset \\ 1895 Q Train\\ 52 Q Test\\ 25 Q Development \\Tafseer\\TYDI QA};
    \node[box, right=of NewDS] (Language Models) {LM};
    \node[circlebox, below=1cm of Language Models] (Final) {FineTuning};
    \node[circlebox, below=1cm of Final] (Quality) {Question};
    \node[circlebox,right=1cm of Quality] (Answer) {Answer};

    \draw[arrow] (Language Models) -- node[right] {Samples} (Final);
    \draw[arrow] (QA) -- node[below] {DS Manipulation} (NewDS);
    \draw[arrow] (NewDS) -- (Language Models);
    \draw[arrow] (Final) -- (Quality);
    \draw[arrow] (Final) -- (Quality);
    \draw[arrow] (Final) -- (Answer);
    \draw[arrow] (Quality) -- (Answer);

    \end{tikzpicture}
    \caption{Architecture Diagram: The workflow for dataset expansion and model fine-tuning. The old dataset is manipulated to create a larger set, which is then fed into various Language Modelss for fine-tuning, resulting in improved question-answer pairs.}
    \label{fig:architecture-diagram}
\end{figure}

This section details the structured process followed to create and improve the dataset utilized in Qur’anic Question Answering (QA) systems. The methodology includes key phases: data collection, dataset expansion, data cleaning, and model fine-tuning. Each phase is thoroughly designed to ensure the quality, reliability, and efficacy of the dataset, establishing a strong foundation for robust QA performance (see Figure \ref{fig:architecture-diagram}).

\subsection{Data Collection}
\label{sec:datacollection}

The dataset was collected from various trustworthy sources to provide a wide range of questions and related Qur’anic verses. The aim was to gather reliable data of high quality to increase the dataset’s scope and improve the QA model’s capability to handle diverse question-answer pairs. The initial dataset was expanded significantly by integrating data from the following sources:

\begin{itemize}
    \item \textbf{Quran QA 2022 Dataset:}  This foundational dataset, available on GitHub and curated by Mohammed Elkomy, was used for initial experiments \cite{elkomy2022quran}. It features structured questions and annotated Qur’anic passages, focusing on both factoid and non-factoid queries.
    \item \textbf{Kaggle Dataset:} The”Quran QA” dataset from Kaggle provided additional questions and corresponding passages, enriching the dataset’s diversity \cite{mobassir2024kaggle}.
    
    \item \textbf{Tafseer Book PDF:} the new dataset was increased by using 1000 Questions and Answers in the Holy Quran, a Tafseer-based resource. Relevant question-passage pairs were meticulously extracted and cleaned to ensure high-quality integration \cite{ashor2023noor}.
    
    \item \textbf{Hugging Face Datasets:}: two key datasets were used from Hugging Face: 
    \begin{itemize}
        \item \textit{Quran-TafseerBook Dataset} by Mohamed Rashad, featuring classical Tafseer texts \cite{rashad2024tafsir}.
        \item \textit{Quran-Classical-Arabic-English Parallel Texts} by ImruQays, offering parallel translations
        for enriched linguistic context \cite{imruqays2024parallel}.
    \end{itemize}
    \item \textbf{List of Plants Citation in Quran and Hadith:} This resource from the Qur’anic Botanic Garden provided context-specific references to plants mentioned in the Qur'an and Hadith, adding another dimension to the dataset \cite{plants2024citation}.
\end{itemize}

\subsection{Dataset Expansion}

To enhance the dataset, the original dataset was manipulated 251-question dataset by rephrasing and generating additional questions, ultimately expanding it to 629 questions. All of these questions were rephrased twice, resulting in a robust dataset of 1895 questions, which categorized into single-answer, multi-answer, and zero-answer types. This process is illustrated in Figure 1, which shows the transformation of the old dataset through” DS Manipulation” into a more comprehensive version that feeds into language models (Language Models) for fine- tuning. The expanded dataset enabled us to adjust multiple pre-trained transformers models, with a focus on enhancing performance on Qur’anic passage retrieval \cite{qamar2024benchmark}. By varying the phrasing, the dataset’s ability was improved by train models that are more flexible and effective in handling different question formats and vocabulary, ultimately enhancing generalization and performance in QA tasks.

\begin{figure}
    \centering
    \includegraphics[width=1\linewidth]{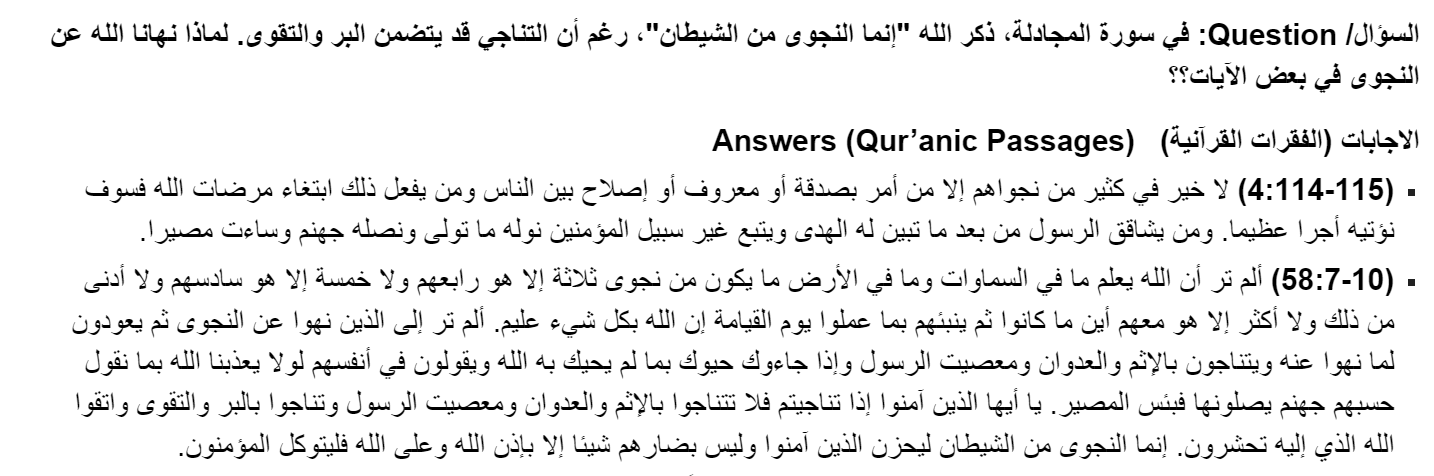}
    \caption{A sample from shared task A. We retrieve the most relevant Qur’anic segment}
    \label{fig:enter-label}
\end{figure}

\subsection{Data Cleaning}

A rigorous data cleaning process was applied to ensure consistency and reliability:

\begin{itemize}
    \item \textbf{Duplicate Removal:}  All duplicate questions and passages were identified and eliminated to maintain dataset integrity.
    \item \textbf{Formatting Standardization:} The text was normalized to make sure that queries were in Modern Standard Arabic (MSA) and Qur’anic verses were in Classical Arabic. This consistency makes model training more effective and enhances semantic matching.
\end{itemize}

\subsection{Language Model Fine-Tuning}
Fine-tuning pre-trained language models is a essential element of this study, as it greatly improves the model’s capability to correctly recognize and understand Qur’anic verses when answering questions in Modern Standard Arabic (MSA). Several advanced transformer models have been modified to address the complex language structure and semantic complexity of the `Quran. the models were enhanced using a larger dataset of 1,895 different questions. This dataset is crafted to encompass diverse linguistic expressions while preserving the intended   meaning of each question.
A diverse set of models was selected for fine-tuning, each offering unique strengths that contribute to overall performance. These models include AraBERT-base, AraBERT-large, CAMeLBERT, AraELECTRA, Roberta-base, and BERT, all of which have demonstrated efficacy in processing Arabic language tasks but required careful adaptation to the context of Qur’anic QA. Here, fine-tuning process was described and the specific modifications for each model:
\begin{enumerate}
    \item AraBERT-base and AraBERT-large \cite{antoun2020arabert} are transformer-based models that have been pre-trained in a large Arabic text corpus. The Ara-BERT base model is characterized by 12 layers and 768 hidden units, while the Ara-BERT large model has a more complex architecture of 24 layers and 1024 hidden units. Both models were fine-tuned using the enhanced dataset, with an emphasis on improving their capacity to manage intricate sentence structures and to grasp the profound meanings embedded in Qur’anic verses. Notably, AraBERT-large showed significant advancements in semantic matching and in identifying instances where there is” no answer” largely due to the larger dataset utilized in its training.
    \item CAMeLBERT \cite{inoue2021camelbert}: CAMeLBERT is another Arabic-specific language model that has been trained on a balanced dataset covering diverse Arabic dialects and MSA. CAMeLBERT was finetuned to leverage its robust language understanding capabilities, especially for questions requiring context sensitive retrieval. The model was further optimized to address the stylistic and grammatical differences between MSA and Classical Arabic, which are prevalent in the Qur’an.
    \item AraELECTRA \cite{antoun2021araelectra}: AraELECTRA employs a discriminative pre-training
    method, focusing on distinguishing real tokens from replaced ones. This model has shown particular strength in tasks
    requiring a fine-grained understanding of token-level semantics. During fine-tuning, AraELECTRA was adopted to new dataset,
    enhancing its ability to accurately detect the presence or absence of relevant verses and improve performance on questions where 
    subtle word choices determine the answer.
    \item Roberta-base \cite{liu2019roberta}: Although Roberta-base is a general-purpose transformer model, it has been adapted for Arabic NLP tasks through extensive pre-training on large texts. Roberta-base was finetuned to improve its passage retrieval capabilities, focusing on optimizing attention mechanisms to better capture relationships between questions and Qur’anic text. The model’s effectiveness was optimized in accurately ranking relevant verses by optimizing hyperparameters and adjusting training procedures.
    \item BERT \cite{devlin2019bert}: The BERT model serves as a foundational architecture for many NLP tasks. A finetuned version of BERT that had previously been trained on the Squad dataset and then readapted it for Qur’anic question answering. This step involved focusing on span prediction capabilities, which are critical for extracting precise answers from passages. Fine-tuning BERT on  enriched dataset allowed to leverage its deep contextual understanding to handle questions with complex or ambiguous phrasing. 
\end{enumerate}

Overall, the fine-tuning process involved extensive experimentation with different hyperparameters, including learning rates, batch sizes, and training epochs, to achieve optimal performance. In this work intensive evaluation procedures were implemented to ensure that each model was capable of not only retrieving accurate answers but also recognizing cases where no relevant answer existed in the Quran (zero answer questions). Additionally, various techniques were integrated such as dropout regularization and gradient clipping to improve model stability and generalization. The outcome of this comprehensive fine-tuning approach is a set of highly specialized models that are well-equipped to handle the linguistic and contextual challenges inherent in Qur’anic QA.

\section{Experimental Design}

This paper experimental design involves fine-tuning several pre-trained Arabic Language Models to tackle the tasks of Question Answering (QA) and Passage Retrieval (PR) in the context of Qur’anic texts. These models, including AraBERT, CAMeLBERT, BERT, Roberta, and AraELECTRA, have been adapted to better address the unique linguistic and semantic challenges presented by the Qur’an. The fine-tuning process, utilizes expanded dataset of 1895 questions derived from an initial set of 251 questions.

Transfer learning was employed to enhance model performance, leveraging the pre-trained knowledge of each model and adapting it through a targeted finetuning process. Moreover, implement ensemble learning techniques, combining predictions from multiple fine-tuned models. This approach improves overall answer accuracy and robustness by leveraging the strengths of each model while mitigating individual weaknesses. Furthermore, a thresholding mechanism was used to filter out low-confidence predictions, effectively managing zero-answer cases by discarding uncertain or ambiguous results.
The evaluation framework uses the following metrics to measure model performance:

\begin{itemize}
    \item \textbf{Mean Average Precision (MAP)}: Measures the precision of each rank position and compares it to the average in all queries, allowing a complete ranking quality assessment.
    \item \textbf{Mean Reciprocal Rank (MRR)}: Assesses the position of the first useful responses in the result list and indicates how well the model finds the correct answer.
    \item \textbf{Recall}:Measures such as recall@5 and recall@10 are used to evaluate the effectiveness of the model in capturing relevant information in the highest rankings.
\end{itemize} 
\section{Results and Discussion}

The evaluation results indicate that fine-tuning various pre-trained models significantly enhanced the performance metrics for question answering and passage    retrieval tasks. This section discusses the results presented in Tables 1 ,2 and 3. 

Table 1 compares the Mean Average Precision (MAP@10) and Mean Reciprocal Rank (MRR) measurements of each model before and after adjustment.\textbf{ AraBERT-base} the most notable improvement was MAP@10, which increased from 0.22 to 0.36, and MRR, from 0.37 to 0.59. These findings indicate that the retrieval accuracy and ranking accuracy have improved significantly, and show that the model has improved its ability to effectively capture relevant passages. The \textbf{Roberta} model, moderate improvements (MAP@10 was 0.05 to 0.12, MRR was 0.01 to 0.17), but it benefited from fine tuning. \textbf{AraBERT-large} remained relatively stable, with only slight increases in MAP@10 and MRR, suggesting that the model was already well-tuned for these tasks or required additional modifications for further improvements. \textbf{CAMeLBERT-base} and \textbf{AraELECTRA- base} demonstrated balanced enhancements, with moderate improvements in both metrics. \textbf{BERT-squad-accelerate}, on the other hand, achieved significant progress, with MAP@10 increasing from 0.07 to 0.25 and MRR from 0.12 to 0.40, highlight the importance of dataset expansion and fine-tuning approach.

Table 2 illustrates the recall performance of various models at different cut-off points. AraBERT-base demonstrated a notable improvement in recall metrics, with Recall@5 increasing from 0.25 to 0.37 and recall@100 rising from 0.30 to 0.50. This indicates a stronger capacity for the model to retrieve relevant passages in the top ranks. In contrast, Roberta showed limited enhancements, with Recall@5 improving only from 0.10 to 0.18, suggesting potential challenges in grasping the nuances of Qur’anic text.  The AraBERT-large model exhibited moderate improvements across recall metrics, with Recall@5 moving from 0.31 to 0.34 and maintaining consistent performance in other recall metrics.
CAMeLBERT-base achieved strong results, improving Recall@5 from 0.32 to 0.36 and showing potential in handling more complex queries, while AraELECTRA- base displayed significant gains, especially in Recall@15, with an increase from 0.42 to 0.57. BERT-squad-accelerate maintained stable recall values at lower cut-off points but significantly performed better in handling unanswerable questions, as showed by a No Answer Recall of 0.75, up from 0.25. This highlights its effectiveness in addressing queries without clear answers.
Table 3 presents precision metrics across various cut-off points. AraBERT- base showed significant improvements, with Precision@5 rising from 0.18 to 0.25 and Precision@100 increasing from 0.01 to 0.06. These enhancements underscore the model’s effectiveness in accurately identifying relevant passages. Roberta showed only small improvement in precision, with Precision@5 moving from 0.05 to 0.10, which corresponds with its overall average performance. Both AraBERT- large and CAMeLBERT-base demonstrated improved precision, particularly at higher cut-off points, indicating their ability to deliver more relevant results, with CAMeLBERT-base achieving Precision@5 of 0.27 compared to its base score of 0.23. The AraELECTRA-base model displayed balanced precision gains, showing   a notable increase from 0.21 to 0.30 at Precision@5. 

Meanwhile, BERT-squad- accelerate excelled in handling no-answer scenarios, attaining a No Answer Precision of 0.75, significantly higher than its base score of 0.25, highlighting its strength in addressing unanswerable questions.

Overall, the results underscore the importance of model architecture and dataset expansion in enhancing the performance of QA systems for Holy Qur’an. The use of ensemble learning and thresholding mechanisms contributed to   the robustness of using these models, particularly in managing zero-answer cases. Future work could explore optimizing specific model architectures further and integrating additional data sources to enhance semantic understanding and retrieval performance.
\begin{table}[H]
\centering
\caption{Comparison of multiple model versions based on MAP@10 and MRR evaluation metrics.}
\footnotesize
\setlength{\tabcolsep}{2pt} 
\begin{tabular}{lcc|cc}
\toprule
\textbf{Model} & \multicolumn{2}{c}{\textbf{MAP@10}} & \multicolumn{2}{c}{\textbf{MRR}} \\ 
\cmidrule(lr){2-3} \cmidrule(lr){4-5}
& \textbf{Baseline} & \textbf{Ours} & \textbf{Baseline} & \textbf{Ours} \\ 
\midrule
\textbf{AraBERT-base}    & 0.22 & \textbf{0.36} & 0.37 & \textbf{0.59} \\ 
\textbf{Roberta}         & 0.05 & \textbf{0.12} & 0.01 & \textbf{0.17} \\ 
\textbf{AraBERT-large}   & 0.28 & \textbf{0.28} & 0.40 & \textbf{0.42} \\ 
\textbf{Camelbert-Base}  & 0.32 & \textbf{0.34} & 0.45 & \textbf{0.47} \\ 
\textbf{Araelectra-Base} & 0.21 & \textbf{0.33} & 0.29 & \textbf{0.45} \\ 
\textbf{bert-squad-accelerate} & 0.07 & \textbf{0.25} & 0.12 & \textbf{0.40} \\ 
\bottomrule
\end{tabular}
\end{table}

\begin{table}[H]
\centering
\caption{Comparison of multiple model versions based on Recall evaluation metrics.}
\footnotesize
\setlength{\tabcolsep}{0.8pt}

\begin{tabular}{lcc|cc|cc|cc|cc}
\toprule
\textbf{Overall Recall} & \multicolumn{2}{c}{\textbf{R@5}} & \multicolumn{2}{c}{\textbf{R@10}} & \multicolumn{2}{c}{\textbf{R@15}} & \multicolumn{2}{c}{\textbf{R@100}} & \multicolumn{2}{c}{\textbf{No Answer}}  \\ 
\cmidrule(lr){2-3} \cmidrule(lr){4-5} \cmidrule(lr){6-7} \cmidrule(lr){8-9} \cmidrule(lr){10-11}
& \textbf{Base} & \textbf{Ours} & \textbf{Base} & \textbf{Ours} & \textbf{Base} & \textbf{Ours} & \textbf{Base} & \textbf{Ours} & \textbf{Base} & \textbf{Ours}  \\ 
\midrule
\textbf{AraBERT-base}          & 0.25 & \textbf{0.37} & 0.30 & \textbf{0.50} & 0.30 & \textbf{0.50} & 0.30 & \textbf{0.50}  & 0.00 & \textbf{0.25}  \\ 
\textbf{Roberta}               & 0.10 & \textbf{0.18} & 0.14 & \textbf{0.30} & 0.14 & \textbf{0.30} & 0.14 & \textbf{0.30} & 0.00 & \textbf{0.25}  \\ 
\textbf{AraBERT-large}         & 0.31 & \textbf{0.34} & 0.38 & \textbf{0.40} & 0.38 & \textbf{0.40} & 0.38 & \textbf{0.40} & 0.00 & \textbf{0.25}  \\ 
\textbf{Camelbert-Base}        & 0.32 & \textbf{0.36} & 0.46 & 0.40 & 0.46 & 0.40 & 0.46 & 0.40 & 0.25 & \textbf{0.50}  \\ 
\textbf{Araelectra-Base}       & 0.31 & \textbf{0.49} & 0.42 & \textbf{0.57} & 0.42 & \textbf{0.57} & 0.42 & \textbf{0.57} & 0.25 & \textbf{0.50}  \\ 
\textbf{bert-squad-accelerate} & 0.07 & \textbf{0.32} & 0.09 & \textbf{0.35} & 0.09 & \textbf{0.35} & 0.09 & \textbf{0.35} & 0.25 & \textbf{0.75}  \\ 
\bottomrule
\end{tabular}
\end{table}

\begin{table}[H]
\centering
\caption{Comparison of multiple model versions based on Precision evaluation metrics.}
\footnotesize
\setlength{\tabcolsep}{0.8pt}
\begin{tabular}{lcc|cc|cc|cc|cc}
\toprule
\textbf{Precision} & \multicolumn{2}{c}{\textbf{P@5}} & \multicolumn{2}{c}{\textbf{P@10}} & \multicolumn{2}{c}{\textbf{P@15}} & \multicolumn{2}{c}{\textbf{P@100}} & \multicolumn{2}{c}{\textbf{No Answer}}  \\ 
\cmidrule(lr){2-3} \cmidrule(lr){4-5} \cmidrule(lr){6-7} \cmidrule(lr){8-9} \cmidrule(lr){10-11}
& \textbf{Base} & \textbf{Ours} & \textbf{Base} & \textbf{Ours} & \textbf{Base} & \textbf{Ours} & \textbf{Base} & \textbf{Ours} & \textbf{Base} & \textbf{Ours}  \\ 
\midrule
\textbf{AraBERT-base}          & 0.18 & \textbf{0.25} & 0.12 & \textbf{0.20} & 0.08 & \textbf{0.15} & 0.01 & \textbf{0.06}  & 0.00 & \textbf{0.25}  \\ 
\textbf{Roberta}               & 0.05 & \textbf{0.10} & 0.04 & \textbf{0.10} & 0.02 & \textbf{0.08} & 0.01 & \textbf{0.05}  & 0.00 & \textbf{0.25}  \\ 
\textbf{AraBERT-large}         & 0.18 & \textbf{0.20} & 0.13 & \textbf{0.15} & 0.09 & \textbf{0.12} & 0.01 & \textbf{0.05}  & 0.00 & \textbf{0.25}  \\ 
\textbf{Camelbert-Base}        & \textbf{0.23} & 0.27 & \textbf{0.19} & 0.20 & \textbf{0.14} & 0.16 & 0.05 & \textbf{0.09}  & 0.25 & \textbf{0.50}  \\ 
\textbf{Araelectra-Base}       & 0.21 & \textbf{0.30} & 0.18 & \textbf{0.24} & 0.13 & \textbf{0.18} & 0.05 & \textbf{0.10}  & 0.25 & \textbf{0.50} \\ 
\textbf{bert-squad-accelerate} & 0.08 & \textbf{0.22} & 0.06 & \textbf{0.18} & 0.05 & \textbf{0.17} & 0.04 & \textbf{0.13} & 0.25 & \textbf{0.75} \\ 
\bottomrule 
\end{tabular}  
\end{table}

\section{Conclusion}
This study presents an effective approach to improving Qur’anic passage retrieval in question-answering systems by fine-tuning pre-trained Arabic Language Models (LMs) on an expanded dataset of 1,895 questions. Models like AraBERT, CAMeLBERT,Roberta , AraELECTRA, and  BERT  ,optimized  using transfer learning, showed improved ability to realize the difficulties of both the Qur’anic text and user queries. Ensemble learning further boosted accuracy and developments, while a thresholding mechanism ensured reliable answers and managed zero-answer cases. The paper results highlight the significance of thorough datasets and sophisticated model structures in creating strong QA systems for Holy Qur’an, contributing both to Natural Language Processing research and providing a valuable resource for Muslims. Future work is recommended to explore additional data sources and further architectural refinements to address challenges like semantic understanding and unanswerable queries (zero answers).
 
\section{Acknowledgment}
Heartfelt gratitude is extended to AiTech AU, \textit{AiTech for Artificial Intelligence and Software Development} (\url{https://aitech.net.au}), for funding this research, providing technical support, and enabling its successful completion.

\begin{flushleft}

\end{flushleft}

\end{sloppypar}
\end{document}